\pdfoutput=1

\documentclass[11pt]{article}

\usepackage[]{ACL2023}

\usepackage{times}
\usepackage{latexsym}
\usepackage{hyperref}
\usepackage{multirow}
\usepackage[T1]{fontenc}

\usepackage[utf8]{inputenc}
\usepackage{graphicx}
\usepackage{microtype}

\usepackage{inconsolata}

\usepackage{ctable} 

\title{My Boli: Code-mixed Marathi-English Corpora, Pretrained Language Models and Evaluation Benchmarks}

\author{Tanmay Chavan$^1$\thanks{~~First author, equal contribution}~~, 
        Omkar Gokhale$^1$$\footnotemark[1]$~~,
        Aditya Kane$^1$$\footnotemark[1]$~~,
        Shantanu Patankar$^1$$\footnotemark[1]$~~,
        Raviraj Joshi$^2$\\
  Pune Institute of Computer Technology, L3Cube$^1$\\
  Indian Institute of Technology Madras, L3Cube$^2$\\
  \texttt{\{chavantanmay1402}, \texttt{omkargokhale2001}, \texttt{adityakane1}, \texttt{shantanupatankar2001\}@gmail.com},\\ 
  \texttt{ravirajoshi@gmail.com} \\
}

\begin{document}
\maketitle
\begin{abstract}

The research on code-mixed data is limited due to the unavailability of dedicated code-mixed datasets and pre-trained language models. In this work, we focus on the low-resource Indian language Marathi which lacks any prior work in code-mixing. We present L3Cube-MeCorpus, a large code-mixed Marathi-English (Mr-En) corpus with 10 million social media sentences for pretraining. We also release L3Cube-MeBERT and MeRoBERTa, code-mixed BERT-based transformer models pre-trained on MeCorpus. Furthermore, for benchmarking, we present three supervised datasets MeHate, MeSent, and MeLID for downstream tasks like code-mixed Mr-En hate speech detection, sentiment analysis, and language identification respectively. These evaluation datasets individually consist of manually annotated \url{~}12,000 Marathi-English code-mixed tweets. Ablations show that the models trained on this novel corpus significantly outperform the existing state-of-the-art BERT models. This is the first work that presents artifacts for code-mixed Marathi research. All datasets and models are publicly released at \url{https://github.com/l3cube-pune/MarathiNLP} . 

\end{abstract}

\section{Introduction} 

The modern world has been engulfed by the presence of social media platforms like Twitter and Facebook \cite{impactoftwitter}. Moreover, websites like YouTube have witnessed considerable user interaction in the comments section of videos \cite{youtubecomments}. These posts and comments closely reflect the thoughts of the general public. It is common amongst users to discuss social, political, and other topics over such social media. This leads to users using a mixed language for communicating over social media platforms.

Code-mixing is known as the mixing of words from multiple languages while retaining the script of a single language. Most commonly, the Latin script is used to encapsulate the words of multiple languages. For example, a given text can be of the Marathi language written in Latin script, as opposed to the Devnagari script, which is the original script of the Marathi language \cite{joshi2022l3cube_mahacorpus}. Code-mixed data is inherently difficult to process and analyze due to its linguistic complexity, variance in spelling and grammar, and long-tailed distribution of uncommon terms and phrases, which are often specific to the geography and demographic of the source location. It is observed that a large number of tweets, comments, and posts on social media are code-mixed in nature. Thus, with the advent of social media analytics, effectively analyzing code-mixed data has gained the utmost importance.

Marathi is a language which has its origins in Maharashtra, a state in India. Due to the state's geographic and demographic expanse, Marathi has evolved into a language with multiple varieties and dialects. Recently, there has been some focus on Marathi NLP based on the Devanagari script \cite{joshi2022l3cube_mahanlp,joshi2022l3cube_mahacorpus,kulkarni2021l3cubemahasent,patil2022l3cube,litake2022l3cube}. However, a large chunk of tweets, posts, and comments in Marathi are in code-mixed form. In spite of this, no efforts have been made to curate models and datasets pertaining to Marathi code-mixed data in the past. This work presents the following.
\begin{enumerate}
    \item We release three supervised datasets (L3Cube-MeHate, MeLID, and MeSent) and one unsupervised dataset (L3Cube-MeCorpus)\footnote{\href{https://github.com/l3cube-pune/MarathiNLP}{MarathiNLP}}. The unsupervised corpus consists of 5 million (70.9M tokens) Roman script code-mixed Marathi-English samples compiled from various sources. We further include 5M Devnagari sentences based on original text making it a 10 million (139.5M tokens) mixed-script MeCorpus.
    \item The supervised dataset contains labels for code-mixed Marathi-English (Roman script) hate classification, sentiment detection, and language identification. These datasets were manually annotated by native Marathi speakers. 
    \item Finally, we release a plethora of code-mixed MeBERT-based pre-trained and fine-tuned models for downstream tasks trained on these novel corpora. These models include MeBERT\footnote{\href{https://huggingface.co/l3cube-pune/me-bert}{MeBERT}}, MeBERT-Mixed\footnote{\href{https://huggingface.co/l3cube-pune/me-bert-mixed}{MeBERT-Mixed}}, MeBERT-Mixed-v2\footnote{\href{https://huggingface.co/l3cube-pune/me-bert-mixed-v2}{MeBERT-Mixed-v2}}, MeRoBERTa-Mixed\footnote{\href{https://huggingface.co/l3cube-pune/me-roberta-mixed}{MeRoBERT-Mixed}}, and MeRoBERTa\footnote{\href{https://huggingface.co/l3cube-pune/me-roberta}{MeRoBERTa}}. The models suffixed as 'Mixed' were trained on full 10M MeCorpus while others were trained on 5M Roman MeCorpus. The supervised models include MeSent-RoBERTa\footnote{\href{https://huggingface.co/l3cube-pune/me-sent-roberta}{MeSent-RoBERTa}}, MeHate-RoBERTa\footnote{\href{https://huggingface.co/l3cube-pune/me-hate-roberta}{MeHate-RoBERTa}}, and MeLID-RoBERTa\footnote{\href{https://huggingface.co/l3cube-pune/me-lid-roberta}{MeLID-RoBERTa}}.  
\end{enumerate}


This work is a major milestone towards democratizing NLP for the Marathi language. Additionally, we present several ablations with fine-tuned models. This is the first work to present a large unsupervised corpus, multiple pre-trained models, and high-quality supervised datasets. This work is a strong foundation in the domain of Marathi and code-mixed Marathi NLP.


\begin{figure*}
    \centering
    \includegraphics[width=\textwidth]{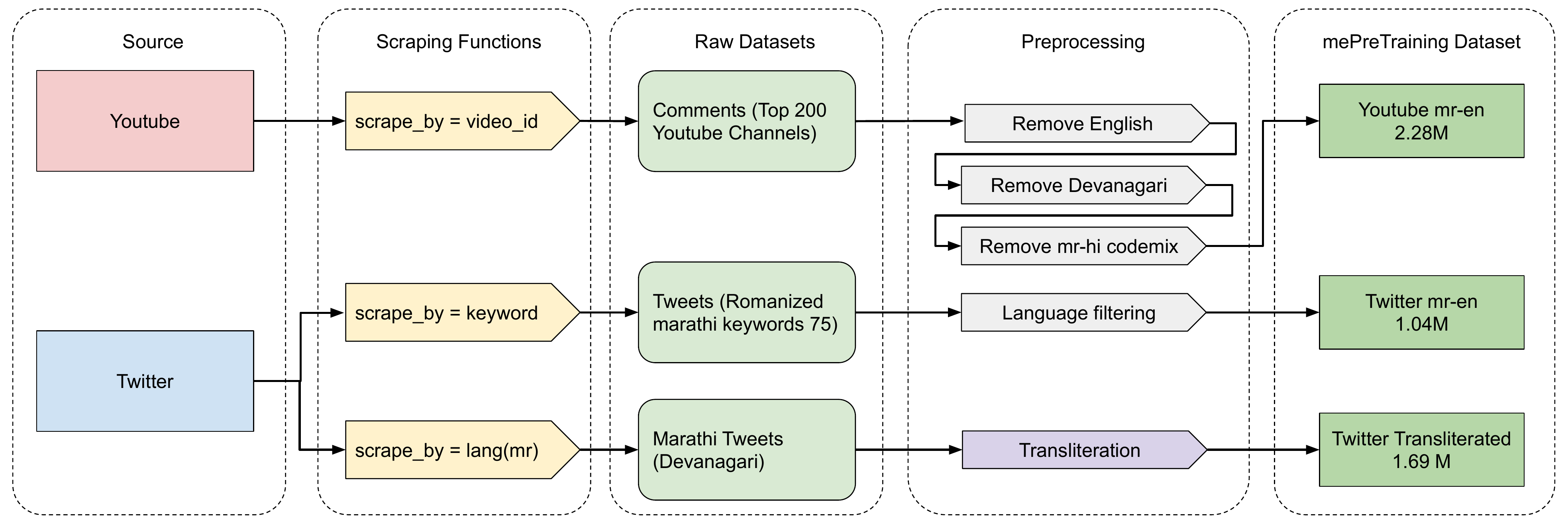}
    \caption{Dataset creation process for our 5 million code-mixed corpora. These 5 million examples are further transliterated to Devnagari script, which results in a combined corpora of 10 million examples.}
    \label{fig:myimage}
\end{figure*}

\section{Related Works}


The use of regional scripts, such as Devanagari, Gurmukhi, Bengali, etc., presents a significant challenge in India due to keyboards primarily designed for the Roman script and the population's familiarity with it. The demand for code-mix datasets and models tailored to regional languages has increased exponentially. These resources play a crucial role in enabling enhanced analysis and moderation of social media content that is code-mixed.

In the realm of language models, BERT-based architectures \cite{vaswani2017attention}, including variations such as RoBERTa \cite{liu2019roberta} and ALBERT \cite{lan2019albert}, have gained popularity due to their application in pre-training and fine-tuning on various tasks. Multilingual models like multilingual-BERT and XLM-RoBERTa 
\cite{conneau2019unsupervised} have specifically focused on data representations that are multilingual and cross-lingual in nature, offering improvements in accuracy and latency. However, these models are pre-trained on 
 less than a hundred thousand real code-mix texts.

While previous research efforts have addressed code mixing in other Indian languages, the specific domain of code-mix Marathi remains largely unexplored. Notably, there is a scarcity of prior work and an absence of a dedicated code-mix Marathi dataset. However, other Indian languages have seen some notable contributions. For instance, \citet{hande2020kancmd} presents KanCMD a code-mixed Kannada dataset for sentiment analysis and offensive language detection. \citet{HASOC-dravidiancodemix-2021} and  have released datasets encompassing Tamil-English and Malayalam-English code-mixed texts. \citet{nayak2022l3cube} have made available HingCorpus, a Hindi-English code-mix dataset and also open-sourced pre-trained models trained on code-mix corpora. \citet{srivastava2021hinge} provide HinGE, a dataset for the generation and evaluation of code-mixed Hinglish text, and demonstrate techniques for algorithmically creating synthetic Hindi code-mixed texts.

In the realm of transliteration, there have been attempts to pre-train language models using transliterated texts. However, these models often underperform due to the rule-based nature of most transliteration techniques, which struggle to account for the diverse spelling variations present in real-life code-mixed texts \cite{santy-etal-2021-bertologicomix}.

\section{MeCorpus - Pretraining Data Creation}
We introduce MeCorpus, a new pre-training corpus of 10 million code-mixed Marathi sentences. This consists of 5M Roman and 5M Devanagari sentences. These sentences are extracted from the social media platforms Youtube and Twitter. We also used synthetic data obtained by transliterating Devanagari's tweets. The complete data collection process is illustrated in Figure \ref{fig:myimage}.

\subsection{Twitter data}
A part of the pretraining corpus is obtained from the social networking site Twitter. We utilize \texttt{snscrape}, a scraper for social networking sites to scrape the data from Twitter. We use a keyword-based approach to curate the data. We use frequently used Marathi words as keywords and fetch all the tweets containing the given word. Proper care is taken to manually check that the word is predominantly exclusive to the Marathi language and doesn't occur in texts from other languages. We fetch a fixed number of tweets belonging to a certain keyword and discard the keyword if the tweets fail to satisfactorily meet the aforementioned conditions after manual verification. Otherwise, we scrape all the tweets containing the keyword and add them to our corpus. This manual verification process ensures that the curated data largely contains Marathi text. 

The Twitter data amounts to over a million tweets. All of the tweets at least partly contain code-mixed Marathi. A significant number of the tweets exhibit code-switching between Marathi and English. A small portion of the tweets also contains code-switched Hindi-Marathi text. We anonymize the data before using it for pre-training. The username mentions are replaced with the '@USER' text. We also remove links and hashtags from the tweets. The Twitter corpus contains 50M tokens.
\subsection{YouTube data}
Youtube comments are an excellent source of Code-mixed Marathi data. We scrape all the comments from 200 marathi youtube channels using the \texttt{youtube-comments-scraper} library. This gave us a mix of English, Devanagari, and code-mixed Marathi sentences. We then removed the Devanagari and English comments to obtain the code-mixed Marathi data. This data was then pre-processed and used in our pre-training dataset. Devanagari words were identified and removed by checking their utf-8 encoding. We remove all comments which have more than 80\% Devanagari words. This gives us comments that are either English or Marathi-English code-mixed. We used a fast text classifier to identify sentences that are English. We remove these sentences and were left with code-mixed Marathi sentences. Thus at the end of both filtering steps, we are left with 2,278,097 of the original 7,599,588 comments.

\subsection{Transliteration} 
We scraped around 1.7 million Marathi Devanagari tweets from Twitter using the snscrape library. We then used the indic-trans Python library to transliterate 1,685,233 of these Devanagari tweets to Code-mixed Marathi and added them to our dataset.

\begin{table}[]
\centering
\begin{tabular}{|c|c|}
\hline
\textbf{Source}       & \textbf{\begin{tabular}[c]{@{}c@{}}Number of\\ Sentences\end{tabular}} \\ \hline
Twitter tweets        & 1,037,659                                                              \\ \hline
Youtube comments      & 2,277,108                                                              \\ \hline
Transliterated-tweets & 1,685,233                                                              \\ \hline
\textbf{Total} & \textbf{5,000,000}                                                              \\ \hline
\end{tabular}
\caption{Source-wise split of the Roman script pretraining corpus (5M).}
\end{table}

\section{Me Corpus - Transliteration}
We created an additional 5 million Devanagari sentences and added them to the corpus. This was done by transliterating the Tweets and Youtube comments mentioned in sections 3.1 and 3.2 respectively to the Devanagari script. We also used the scraped Devanagari tweets mentioned in section 3.3. This gave us a total of 5M Devanagari sentences. These 5M sentences were used to pre-train the multilingual models mentioned in the future sections.

\section{MeEval - Downstream dataset creation}
\label{sec:downstream}
We aim to create a large dataset of code-mix Marathi-English data, annotated with sentiment and hatefulness. In this study, we selected a set of tweets from a larger corpus of 1,037,659 tweets obtained from the social media platform Twitter. Half of the tweets chosen were posted on Twitter before 2013, and the other half were posted after 2013. This helped to provide a more diachronic distribution of tweets, as the number of tweets posted in the past few years far out-number the old tweets. The tweets were selected randomly apart from this criteria. This ensures a realistic representation of the sentiment, hate, and profanity distributions present in the real-world data across the past several years. To annotate the data, four annotators fluent in Marathi, Hindi, and English languages were chosen. The Cohen's Kappa \cite{cohenkappa} for the annotators is 0.86.

The collected data was labeled according to three distinct categories: sentiment, hate, and language identification. We followed a set of guidelines while labeling the data for ensuring the veracity of the annotation. We annotated the data after anonymizing it. This helped remove any bias or knowledge of the entity posting it. We also disregarded any additional information which could be inferred by us based on external context but is not apparent by reading the text by itself. Here, we outline the dataset statistics and annotation procedure. The dataset statistics are described in Table \ref{tab:statistics_tab}.

\subsection{MeSent Dataset}
The code-mixed Marathi-English sentiment data is termed as MeSent Dataset.
Tweets expressing good or heartening emotions such as thankfulness, happiness, applause, and appreciation are labeled positive. Tweets expressing negative or disheartening emotions like strong dissent, disappointment, sorrow, derision, and hate are labeled negative. Plain facts, statements, and simple responses are labeled neutral. If a tweet contains conflicting emotions, the stronger emotion is chosen.
\begin{itemize}
    \item  +1 indicating a \textit{positive} sentiment,
    \item  -1 indicating a \textit{negative} sentiment, and
    \item  0  indicating a \textit{neutral} sentiment.
\end{itemize}
While annotating the data, we removed unsuitable and ambiguous tweets. Finally, we selected 4,000 tweets from each sentiment category, leading to the dataset containing a total of 12,000 tweets.

\subsection{MeHate Dataset}
For the hatefulness annotation, we labeled any tweets expressing strongly negative feelings such as insults, mockery, abuse, intimidation, and threats as \textit{hateful}. Any tweet not containing such hateful content is labeled as \textit{non-hateful}. We use 1 for hateful content and 0 for non-hateful content. The MeHate dataset contains 1384 hateful and 1384 non-hateful tweets, totaling 2768 tweets. We also release the full 12k labeled tweets with the majority of non-hate labels.

\subsection{MeLID dataset}
Additionally, a Language Identification (LID) dataset is created. Each word within the selected tweets is labeled based on its language as \textit{Marathi}, \textit{English}, or \textit{Other}. The \textit{Other} category contains invalid words, words from languages other than English or Marathi, and literals such as numbers and proper nouns. The MeLID dataset contains 11,814 tweets. For all three supervised datasets, we provide a pre-defined train, test, and validation split of 80:10:10.



\section{Models trained on code-mixed MeCorpus}

We train several well-known models on our novel pretraining corpora. In this section, we outline these models and their training details. 

We used pre-trained BERT, RoBERTa, mBERT, MuRIL, and XLM Roberta as the base models and trained them on the novel MeCorpus using the Masked Language Modelling (MLM) objective. For MLM training, we train the models for two epochs at a learning rate of $2e-5$, with a weight decay of $0.01$ and a mask probability of $0.15$.

The monolingual models were pre-trained on the Roman 5M codemixed data mentioned in section 3. While the multilingual models were trained on the full 10M corpus (5M Roman + 5M Devanagari sentences) mentioned in section 4. The models pre-trained on mixed-script corpus are suffixed as 'Mixed'.

The resulting models were named similarly to the original models, prefixed with "\textbf{me}", which stands for \textbf{M}arathi-\textbf{E}nglish. Therefore, the models MeBERT, MeBERT-Mixed, MeBERT-Mixed-v2, MeRoBERTa-Mixed, and MeRoBERTa are the BERT, mBERT, MuRIL, XLM-RoBERTa, and RoBERTa models trained on the MeCorpus respectively. Note that these models are further "fine-tuned" on the MeCorpus using the MLM training objective. 
 
\section{Results}
We fine-tune our MeBERT models on the MeSent, MeHate, and MeLID datasets as mentioned in \autoref{sec:downstream} and test them on the respective test data. The same process is repeated for their base models and a few state-of-the-art Marathi models like Indic-BERT \cite{kakwani2020indicnlpsuite}, Marathi-Tweets-BERT \cite{patankar2022spread}, and  Marathi Code-mixed Abusive MuRIL \cite{das2022data}. The results obtained from this are showcased in \autoref{tab:results_tab}.
 It is observed that MeBERT-Mixed-v2 outperforms all other models on the MeHate evaluation set with an F1 score of 78.3\%. For the sentiment analysis corpus MeSent, MeRoBERTa outperforms the others by obtaining an F1 score of 67.27\%. Testing the models on the MeLID dataset, MeBERT-Mixed-v2 outperforms the other models by obtaining an F1 score of 88.6\%. The newly pre-trained code-mixed MeBERT-based models consistently outperform their base models as well as the state-of-the-art Marathi models.



\begin{table}[]
\centering
\resizebox{0.48\textwidth}{!}{
\begin{tabular}{|l|c|c|c|}
\hline
\textbf{Model} & \textbf{MeHate} & \textbf{MeSent} & \textbf{MeLID} \\
    \hline
        Indic-BERT & 61.62 & 55.29 & 87.37 \\ \hline
        MahaTweets-BERT & 63.18 & 57.59 & 87.38 \\ \hline
        Abusive-MuRIL & 67.69 & - & -  \\ \specialrule{.15em}{0em}{0em} 
        BERT & 61.98 & 59.06 & 87.89 \\ \hline
        MeBERT & 73.78 & 61.92 & 88.01 \\ \specialrule{.12em}{0em}{0em}
        mBERT & 66.80 & 60.25 & 87.64 \\ \hline
        MeBERT-Mixed & 77.39 & 65.73 & 88.25 \\ \specialrule{.12em}{0em}{0em}
        MuRIL & 67.93 & 63.38 & 87.55 \\ \hline
        MeBERT-Mixed-v2 & \textbf{78.3} & 64.23 & \textbf{88.6} \\ \specialrule{.12em}{0em}{0em}
        XLM-RoBERTa & 64.66 & 61.06 & 87.54 \\ \hline
        MeRoBERTa-Mixed & 78.07 & 67.17 & 87.42 \\ \specialrule{.12em}{0em}{0em}
        RoBERTa & 66.10 & 58.86 & 86.46 \\ \hline
        MeRoBERTa & 77.85  & \textbf{67.27} & 88.41 \\ \specialrule{.12em}{0em}{0em}

\end{tabular}}
\caption{Macro F1 scores (in \%) of models on the MeHate, MeSent, and MeLID datasets.}
\label{tab:results_tab}
\end{table}

\section{Conclusion}

This work lays the necessary groundwork for future work on code-mixed Marathi. We introduce a novel pretraining corpus of 5 million code-mixed tweets. In addition to that, we present five new models trained on this code-mixed corpus. Furthermore, we present three supervised datasets of 12,000 tweets for hate classification, sentiment analysis, and language identification annotated by native Marathi speakers. We also present thorough ablations and show that our code-mixed MeBERT models outperform the previous state-of-the-art models by a considerable margin.

\section*{Limitations}
A major problem while dealing with Romanized Marathi is the lack of a singular correct spelling of words. A Marathi word can be written in several ways in Marathi, all of which are equally valid and correctly convey meaning despite having significantly different spellings. Developing efficient approaches to tackle this issue will lead to a significant increase in performance on NLP tasks dealing with code-mixed languages. Our keyword-based scraping method uses words primarily from the western Maharashtra dialect of Marathi, which might not sufficiently represent samples from other Marathi dialects. Efforts to increase the dataset to include examples from other dialects will make the dataset more diverse and robust.  

\section*{Ethics Statement}
All of the data used in our experiments has been scraped by legal and valid means, adhering to the provided guidelines. We anonymized the data before usage to protect the privacy of the original authors of the data. This data might contain biases and thus must be used with care. This data also contains strong language which might be unsuitable for some applications. This data should be used only for research purposes and not for training any model for deployment. 

\section*{Acknowledgments}
This work was done under the L3Cube Pune mentorship
program. We would like to express our gratitude towards
our mentors at L3Cube for their continuous support and
encouragement. This work is a part of the L3Cube-MahaNLP project \cite{joshi2022l3cube_mahanlp}.

\bibliography{main}
\bibliographystyle{acl_natbib}
\appendix
\newpage
\section{Appendix}
\label{sec:appendix}


\begin{table}[h]
\begin{tabular}{|c|c|c|c|c|}
\hline
\textbf{Dataset}                & \textbf{\begin{tabular}[c]{@{}c@{}}Sample \\ Count\end{tabular}} & \textbf{\begin{tabular}[c]{@{}c@{}}Avg \\ word \\ count\end{tabular}} & \textbf{Labels} & \textbf{\begin{tabular}[c]{@{}c@{}}Count \\ per \\ label\end{tabular}} \\ \hline
\multirow{3}{*}{\textbf{MeLID}}  & \multirow{3}{*}{11,813}                                             & \multirow{3}{*}{12}                                                        & Marathi         & 99,230                                                                 \\ \cline{4-5} 
                                 &                                                                    &                                                                            & English         & 26,290                                                                 \\ \cline{4-5} 
                                 &                                                                    &                                                                            & Other           & 10,870                                                                 \\ \hline
\multirow{3}{*}{\textbf{MeSent}} & \multirow{3}{*}{12,000}                                            & \multirow{3}{*}{16}                                                        & Positive        & 4,000                                                                  \\ \cline{4-5} 
                                 &                                                                    &                                                                            & Neutral         & 4,000                                                                  \\ \cline{4-5} 
                                 &                                                                    &                                                                            & Negative        & 4,000                                                                  \\ \hline
\multirow{2}{*}{\textbf{MeHate}} & \multirow{2}{*}{2,768}                                             & \multirow{2}{*}{17}                                                        & Non Hate        & 1,384                                                                  \\ \cline{4-5} 
                                 &                                                                    &                                                                            & Hate            & 1,384                                                                  \\ \hline
\end{tabular}
\caption{Statistics for benchmark datasets}
\label{tab:statistics_tab}
\end{table}

\end{document}